\documentclass[runningheads]{llncs}

\usepackage[mobile]{eccv}

\usepackage{eccvabbrv}

\usepackage{graphicx}
\usepackage{booktabs}
\usepackage{comment}
\usepackage{colortbl}
\definecolor{grey}{rgb}{0.5,0.5,0.5}
\usepackage{textcomp}
\usepackage{subcaption}
\usepackage{pifont}
\usepackage{array}
\usepackage{adjustbox}
\usepackage{subcaption}

\usepackage[accsupp]{axessibility}  

\usepackage{hyperref}
\usepackage{orcidlink}

\begin{document}

\title{Geometry-Aware Semantic Reasoning for Training Free Video Anomaly Detection}

\titlerunning{MMVAD}

\author{Ali Zia$^{\ast}$\inst{1} \and
Usman Ali$^{\ast}$\inst{2} \and
Muhammad Umer Ramzan \inst{2}\and
Hamza Abid \inst{2} \and
 Abdul Rehman \inst{2} \and
Wei Xiang\inst{1}}

\authorrunning{Ali et al.}

\institute{School of Computing, Engineering \& Mathematical Sciences,  La Trobe University, Melbourne, Australia \and
School of Engineering and Applied Sciences, GIFT University, Gujranwala 52250, Pakistan
\\}

\maketitle
\begingroup
\renewcommand\thefootnote{$\ast$}
\footnotetext{These authors contributed equally.}
\endgroup

\begin{abstract}
Training-free video anomaly detection (VAD) has recently emerged as a scalable alternative to supervised approaches, yet existing methods largely rely on static prompting and geometry-agnostic feature fusion. As a result, anomaly inference is often reduced to shallow similarity matching over Euclidean embeddings, leading to unstable predictions and limited interpretability, especially in complex or hierarchically structured scenes.
We introduce MM-VAD, a geometry-aware semantic reasoning framework for training free VAD that reframes anomaly detection as adaptive test-time inference rather than fixed feature comparison. Our approach projects caption-derived scene representations into hyperbolic space to better preserve hierarchical structure, and performs anomaly assessment through an adaptive question–answering process over a frozen large language model. A lightweight, learnable prompt is optimised at test time using an unsupervised confidence–sparsity objective, enabling context-specific calibration without updating any backbone parameters. To further ground semantic predictions in visual evidence, we incorporate a covariance-aware Mahalanobis refinement that stabilises cross-modal alignment.
Across four benchmarks, MM-VAD consistently improves over prior training-free methods, achieving 90.03\% AUC on XD-Violence and 83.24\%, 96.95\%, and 98.81\% on UCF-Crime, ShanghaiTech, and UCSD Ped2, respectively. Our results demonstrate that geometry-aware representation and adaptive semantic calibration provide a principled and effective alternative to static Euclidean matching in training-free VAD.


\keywords{Training-free VAD \and Geometry-Aware Semantic Reasoning \and Hyperbolic Space Representation \and Adaptive Test-Time Inference }
\end{abstract}

\section{Introduction}
\label{sec:intro}

\begin{figure}[t]
    \centering
\includegraphics[width=0.6\textwidth]{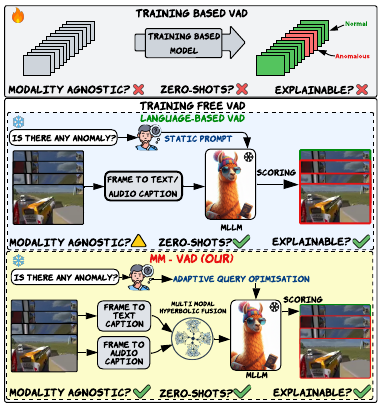}
\caption{Comparison of prior VAD architectures and our modality-agnostic MM-VAD, which introduces adaptive query optimisation and hyperbolic fusion for zero-shot, explainable anomaly detection.}
    \label{fig:problem}
\end{figure}

Video Anomaly Detection (VAD) is a central problem in computer vision with far-reaching applications in surveillance, security, and public safety \cite{b2}. The majority of existing VAD frameworks can be categorised as supervised \cite{b16}, weakly supervised \cite{b3}, or unsupervised \cite{b4}, but all are fundamentally limited by their dependency on extensive labelled datasets or strong assumptions about the normality distribution. Supervised and weakly supervised approaches incur high annotation costs and often struggle to generalise across domains, while unsupervised methods may falter in complex, non-stationary environments due to their fragile modelling of “normal” behaviours \cite{us_intr_1, us_intr_2, us_intr_3, us_intr_4}. Additionally, the computational and operational overhead of retraining these models for each new deployment remains a major barrier to practical adoption. Figure~\ref{fig:problem} illustrates these limitations and contrasts them with our proposed training-free reasoning architecture. Recent progress in foundation models has opened the door to training-free anomaly detection, yet current systems still fail to capture the full multimodal structure of real scenes. Many methods remain purely vision-centric, overlooking anomalies with strong acoustic signatures, as in \textbf{LAVAD}~\cite{b1}. When audio is incorporated, fusion is typically performed in a fixed Euclidean space, as in \textbf{MCANet}~\cite{dev2024mcanet}, which struggles to represent hierarchical cross-modal relationships and cannot generalise when a modality is absent. In parallel, prompting strategies remain largely static, retrieval-based caption refinement and hand-crafted prompts used in \textbf{LAVAD}~\cite{b1}, \textbf{Flashback-IB}~\cite{lee2025}, and \textbf{EventVAD}~\cite{event}, limit semantic flexibility and restrict the depth of reasoning these models can perform. Consequently, existing training-free VAD systems offer little in terms of structured or interpretable explanations, producing anomaly scores without transparent semantic justifications. 
Training-free VAD systems therefore struggle because anomaly reasoning requires hierarchical semantic structure, yet most existing approaches operate in flat Euclidean embedding spaces. We address this limitation by introducing geometry-aware semantic reasoning, where multimodal scene descriptions are represented in hyperbolic space and analysed through adaptive question–answering.

To address these interlinked issues, we introduce a training-free multimodal VAD framework that unifies modality-agnostic processing, hierarchical representation, and explicit reasoning. Our method integrates audio and visual captions in hyperbolic space to model non-linear, multi-level structure and uses an adaptive question–answer mechanism to interpret scene context rather than simply score embeddings. This design enables the model to generate anomaly decisions that are both more stable and more interpretable, explaining what happened and why it departs from normal behaviour. Our work uses “zero-shot” in the standard sense adopted by training-free VAD systems such as LAVAD~\cite{b1}, and Flashback-IB~\cite{lee2025}. All backbone components, including the captioning models, ImageBind encoders, hyperbolic projection layers, and the LLM, remain strictly frozen, and no anomaly labels or supervised objectives are used during inference.
The lightweight optimisation operates only on a small prompt representation at test time and does not modify any backbone parameters, preserving the training-free nature of the framework while enabling context-sensitive semantic reasoning.
Our work introduces geometry-aware semantic reasoning for training-free video anomaly detection and makes the following contributions:

\begin{itemize}
    \item First, we identify a key limitation of existing training-free VAD systems, i.e. anomaly inference is performed in flat Euclidean embedding spaces that fail to capture the hierarchical structure of multimodal events. We introduce a geometry-aware semantic reasoning framework that represents multimodal scene descriptions in hyperbolic space, enabling structured modelling of complex event relationships.
    
    \item Second, we formulate anomaly detection as an adaptive semantic reasoning process over these hierarchical representations. Instead of relying on static prompt templates, we optimise a continuous query representation at test time using an unsupervised objective, allowing the model to adapt its reasoning to the context of each video without updating backbone parameters.

     \item Third, we introduce a robust multimodal reasoning mechanism that improves the stability of training-free inference. Caption stabilisation via temporal self-alignment reduces noise in vision–language outputs, while a covariance-aware cross-modal refinement stage grounds language-based anomaly predictions in the statistical structure of the visual feature distribution.
     
\end{itemize}
 Together, these components form a unified framework for geometry-aware semantic inference in training-free VAD.

\section{Related Work}
Video Anomaly Detection (VAD) has evolved significantly, progressing from fully supervised~\cite{sp4,b14,b15,b16} and weakly supervised~\cite{b10,b17,b18,b19,b3,b21,b23,b24,b25} frameworks, which rely on exhaustive frame- or clip-level annotation, to one-class~\cite{b26,b27,b28,b29,b30,b31,b32} and unsupervised~\cite{b33,b36,b37,b56,b40,b41,b42} paradigms that aim to reduce labelling costs and enhance scalability. However, \textbf{supervised methods} face prohibitive annotation requirements and poor generalisation to novel anomaly types. \textbf{Weakly supervised} methods, including RTFM~\cite{b23} and CLAWS~\cite{b22}, employ multiple instance learning (MIL) to leverage coarse labels, but remain sensitive to ambiguous event boundaries and suffer under domain shift. \textbf{One-class and unsupervised techniques}, such as SHT~\cite{b33} and GCL~\cite{b4}, model normalcy via generative or reconstruction-based frameworks, but are prone to false positives when rare or evolving normal events occur, limiting their robustness in unconstrained scenarios.\\
The advent of large-scale vision--language models (VLMs) and large language models (LLMs), such as CLIP and BLIP-2, has accelerated a clear shift towards \textbf{zero-shot} and \textbf{training-free} VAD. A pioneering system such as \textbf{LAVAD}~\cite{b1} demonstrated that frozen foundation models can perform anomaly detection without any domain-specific training, enabling scalable, plug-and-play deployment in real-world settings. Building on this foundation, later methods such as \textbf{AnyAnomaly}~\cite{ahn2025anyanomaly} introduced user-defined, text-driven anomaly descriptions, while \textbf{Flashback-IB}~\cite{lee2025} employed a memory-driven retrieval mechanism that constructs an offline pseudo-memory of normal and abnormal events for real-time zero-shot inference. Similarly, \textbf{EventVAD}~\cite{event} leveraged unsupervised event segmentation and multimodal reasoning with large video--language models to achieve strong training-free performance across challenging benchmarks. These training-free paradigms have proven particularly valuable in scenarios where labelled data are scarce or domain conditions change frequently.
Recent efforts have begun to close the gap between unimodal and multimodal training-free approaches. For example, \textbf{MCANet}~\cite{dev2024mcanet} integrates audio and visual modalities through Euclidean-space fusion, demonstrating that auditory information can enhance anomaly recognition when visual cues are ambiguous. However, Euclidean fusion can struggle to capture the complex, hierarchical relations inherent in multimodal embeddings. In contrast, our framework advances this line of work by introducing \textbf{hyperbolic semantic fusion}, which more effectively represents cross-modal structure and semantic depth. Combined with adaptive prompting and an event-aware reasoning module, our model offers a general, efficient, and training-free framework for robust multimodal VAD.

\begin{figure}[t]
    \centering
    \includegraphics[width=0.9\textwidth]{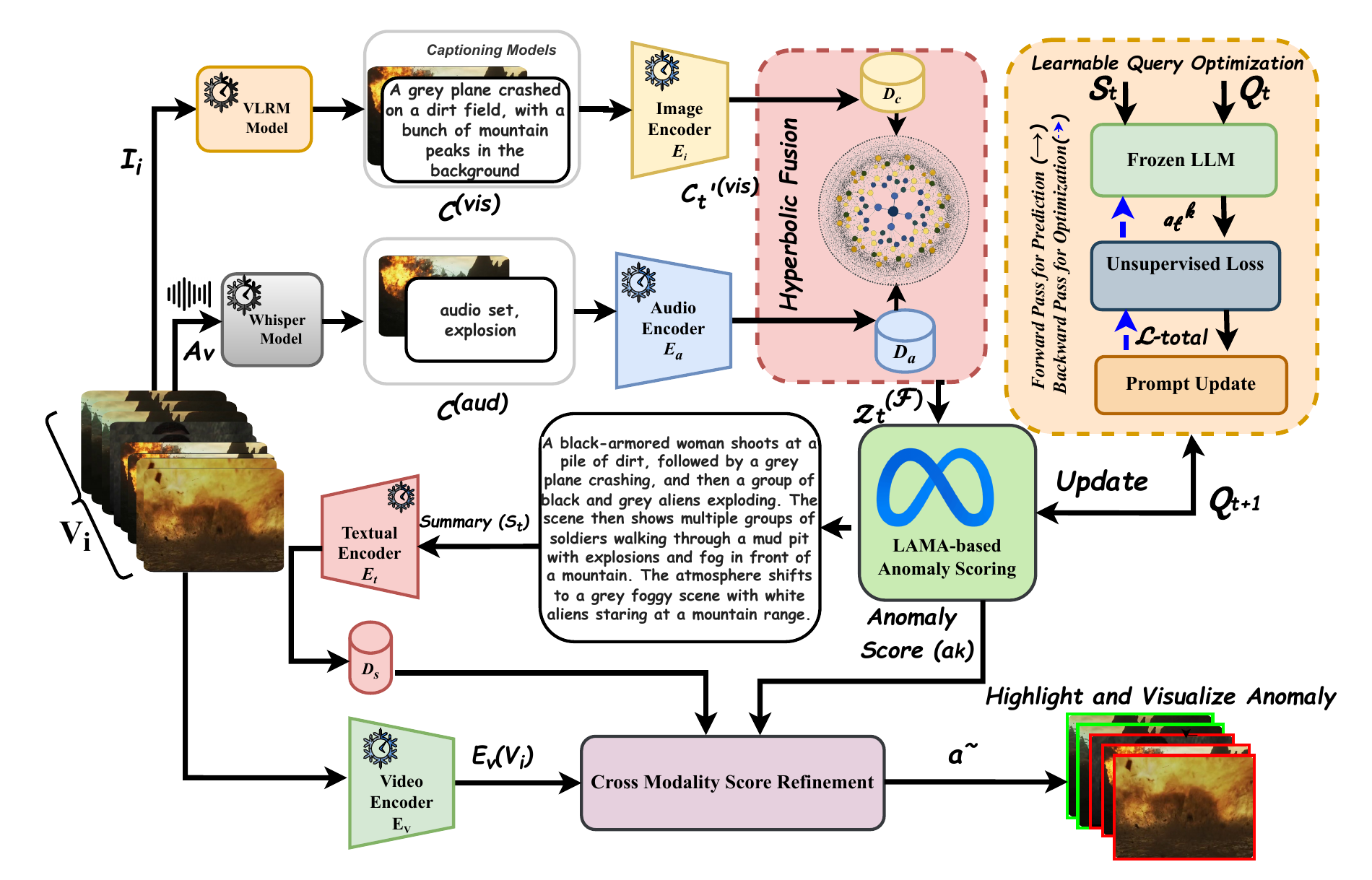}
    \caption{An overview of our MM-VAD framework. Visual captions $C_{\text{vis}}$ and audio captions $C_{\text{aud}}$ are fused in hyperbolic space, and the fused features drive unsupervised refinement of the anomaly query from $Q_t$ to $Q_{t+1}$. A LLaMA-based scorer then evaluates textual video summaries against $Q_{t+1}$, and a cross-modal refinement step produces the final anomaly scores.}
    \label{fig:framework}
\end{figure}

\section{Geometry-Aware Semantic Reasoning for Training-Free VAD}


We propose MM-VAD, a training-free framework for geometry-aware semantic reasoning in video anomaly detection that optionally integrates visual and audio cues within a unified hyperbolic representation space. Hyperbolic embeddings naturally represent hierarchical relationships with exponentially expanding volume, allowing rare or abnormal events to appear as semantic deviations from the dominant normal structure.
The framework is built around a single principle, i.e. geometry-aware semantic reasoning for anomaly detection. Instead of performing anomaly detection through similarity matching in a flat embedding space, MM-VAD models scene semantics as structured representations whose hierarchical relationships are preserved in hyperbolic geometry and interpreted through adaptive language-based reasoning. As illustrated in Figure~\ref{fig:framework}, 
this principle is realised through four components that together implement geometry-aware semantic reasoning. Caption stabilisation enforces semantic consistency, hyperbolic fusion captures hierarchical cross-modal structure, adaptive query optimisation performs context-aware reasoning, and statistical refinement grounds semantic predictions in the visual feature distribution.

\subsection{Problem Formulation}
We reformulate VAD as an adaptive, multimodal question-answering (QA) task. Each video instance is represented as a tuple $(\mathcal{V}, \mathcal{A})$, where $\mathcal{V}=\{I_t\}_{t=1}^N$ is a sequence of $N$ video frames and $\mathcal{A}$ is the corresponding synchronised audio. Our goal is to infer a sequence of anomaly likelihoods $\mathcal{S} = \{s_t\}_{t=1}^N$, with each $s_t \in [0,1]$ indicating the predicted probability of an anomaly at frame $t$.

Crucially, instead of direct classification or reconstruction, we cast VAD as sequential QA over the multimodal scene context. At each time step, a context-rich representation is constructed by fusing visual and auditory captions. To capture complex, hierarchical event semantics, these multimodal embeddings are projected into a shared "hyperbolic space", which models cross-modal relationships and structural dependencies beyond Euclidean fusion. Given this representation, a parameterised, learnable prompt $Q_t$ adaptively queries a frozen LLM, which acts as a semantic reasoner to assess anomalousness at each frame.
Formally, we define a composite scoring function:
\begin{equation}
    \mathcal{F}: (\mathcal{V}, \mathcal{A}) \xrightarrow{\text{captioning, fusion}} \mathcal{Z}^{(H)} \xrightarrow{\text{QA+LLM}} \mathcal{S}
\end{equation}
where $\mathcal{Z}^{(H)}$ denotes the sequence of hyperbolic fused multimodal embeddings, and the final anomaly scores $\mathcal{S}$ are produced by adaptive QA via the LLM, guided by $Q_t$.  

In accordance with the training-free setting, all components are constructed from pre-trained, frozen foundation models $\mathcal{M} = \{\mathcal{M}_j\}$; no task-specific parameter updates are permitted. This framework explicitly departs from conventional VAD, foregrounding adaptive, context-aware reasoning over static, modality-specific scoring.

\subsection{Context-Driven Caption Cleaning}
The initial stage of MM-VAD focuses on generating high-fidelity, semantically rich textual representations from both visual and auditory streams. Given an input video $\mathcal{V}$ and its corresponding audio $\mathcal{A}$, we first perform parallel captioning for each modality. Specifically, for each video segment $I_t$ of fixed size, a Vision-Language Reward Model (VLRM)~\cite{b7} generates a visual caption $\Phi_{\text{vis}}(I_t)$, while the corresponding audio segment $A_t$ is transcribed into text by the Whisper model \cite{radford2023robust}, $\Phi_{\text{aud}}(A_t)$. This yields the following two initial caption sequences:

\begin{equation}
C^{\text{vis}} = \{ \Phi_{\text{vis}}(I_i) \}_{i=1}^N, \qquad C^{\text{aud}} = \{ \Phi_{\text{aud}}(A_i) \}_{i=1}^N
\end{equation}

MM-VAD is explicitly \emph{modality-agnostic} and consumes real audio only when the dataset provides it (e.g., XD-Violence). For video-only benchmarks (e.g., UCF-Crime, ShanghaiTech, UCSD Ped2), the audio branch is disabled and the pipeline runs in a purely visual mode without any retraining or architectural changes. We do \emph{not} synthesise pseudo-audio or hallucinated modalities. In this setting, the hyperbolic fusion layer reduces to unimodal processing. We apply the exponential map to visual caption embeddings and aggregate them via the geodesic mean. All subsequent stages (caption cleaning, adaptive reasoning, score refinement) remain unchanged. As reported in Table~\ref{tab:ucf_xd_combined_grouped}, MM-VAD retains strong performance under this vision-only configuration, which shows that our core contributions (hyperbolic semantic fusion and adaptive prompt-based reasoning) remain effective even when audio is absent.

However, raw captions often contain noise, inconsistencies, or contextually irrelevant information, particularly in unconstrained real-world scenes. To address this, we introduce a context-driven cleaning procedure focused on the visual stream. The core intuition is that in high-frame-rate videos, adjacent small fixed-size video segments frequently depict overlapping or similar content, permitting semantic self-alignment for denoising. For each video segment $I_t$, we search for its most semantically consistent caption within the entire set $C^{\text{vis}}$ by maximising the similarity between the visual embedding $E_{\text{v}}(I_t)$ and the text embeddings $E_{\text{t}}(C_j^{\text{vis}})$ from a frozen vision-language model (ImageBind ~\cite{b63}). The refined caption $C'^{\text{vis}}t$ is then computed as:
\begin{equation}
C'{_t}^{\text{vis}} = \underset{C_j^{\text{vis}} \in C^{\text{vis}}}{\arg\max} ; \text{sim}(E_{\text{v}}(I_t), E_{\text{t}}(C_j^{\text{vis}}))
\end{equation}

\subsection{Hyperbolic Semantic Fusion and Alignment}
To capture the complex, hierarchical relationships inherent in multimodal scene representations, we project the cleaned visual and auditory captions into a shared hyperbolic embedding space. 
This is particularly important for anomaly detection, as many real-world anomalous events involve hierarchical scene structures (e.g., background-foreground interplay, composite events, or temporally extended activities) that are poorly captured by traditional Euclidean representations. Formally, let $C'^{\text{vis}} = \{C_1'^{\text{vis}}, \dots, C_N'^{\text{vis}}\}$ and $C^{\text{aud}} = \{C_1^{\text{aud}}, \dots, C_N^{\text{aud}}\}$ denote the cleaned visual and auditory caption sequences. Each caption is first encoded using a frozen language encoder $E(\cdot)$ (ImageBind ~\cite{b63}), yielding sets of visual and auditory embeddings $z$ in $\mathbb{R}^d$.

To embed these representations in hyperbolic space, we apply a mapping $\mathcal{H}: \mathbb{R}^d \rightarrow \mathbb{H}^d$, where $\mathbb{H}^d$ denotes the $d$-dimensional Poincaré ball model. This mapping can be realised via exponential mapping: 

\begin{equation}
    z_t^{\text{vis}} = \mathcal{H}(E_{i}(C_t'^{\text{vis}})), \qquad
    z_t^{\text{aud}} = \mathcal{H}(E_{a}(C_t^{\text{aud}}))
\end{equation}

Hyperbolic fusion is then performed by aggregating the paired embeddings for each time step $t$. Specifically, we adopt 
a weighted geodesic mean to combine $z_t^{\text{vis}}$ and $z_t^{\text{aud}}$ in $\mathbb{H}^d$:
\begin{equation}
    z_t^{(\mathcal{F})} = \operatorname{GeodesicMean}(z_t^{\text{vis}}, z_t^{\text{aud}})
\end{equation}
where $z_t^{(\mathcal{F})}$ represents the fused multimodal embedding at time $t$.

The key advantage of this fusion is its ability to preserve the underlying hierarchy and compositional structure of events, which is especially valuable for reasoning about anomalies that involve subtle, multi-scale interactions across modalities. Empirically, we find that hyperbolic fusion enhances the discriminability of normal versus anomalous events compared to standard (Euclidean) averaging. The resulting sequence $\{z_t^{(\mathcal{F})}\}_{t=1}^N$ serves as a unified, hierarchically-aware representation of the scene, providing rich semantic context for the subsequent adaptive prompt-based anomaly reasoning stage.

\subsection{Adaptive Prompt-Based Anomaly Reasoning}
\label{subsec:adaptive_prompt}
To exploit the multi-level semantics captured by the hyperbolic fusion module, we cast video anomaly detection as a \emph{context-adaptive question--answering} task over a frozen LLM. The fused hyperbolic embeddings $\mathbf{z}^{(F)}_t$ encode hierarchical relationships across visual and auditory cues; these are converted into structured textual summaries $\mathbf{S}_t$ that expose both global scene context and local event details. The LLM, operating purely at inference time, processes these summaries to infer how typical or atypical each segment is.

A continuous prompt vector $\mathbf{Q}$ serves as a differentiable interface that controls how the LLM interprets the scene summaries. Our aim is to keep the entire process strictly training-free, all backbone encoders and the LLM remain frozen, and no ground-truth labels are used. Rather than relying on manually engineered prompts, we adapt $\mathbf{Q}$ at test time using a fully unsupervised objective that encodes two generic properties of anomaly patterns: predictions should be confident, and anomalies should be rare.

Given a video with $N$ temporal segments, we initialise $\mathbf{Q}_0$ and refine it over $K$ lightweight iterations. At iteration $k$, the current prompt $\mathbf{Q}_k$ is concatenated with each summary $\mathbf{S}_t$ to produce preliminary anomaly scores
\begin{equation}
    a_t^{(k)} = \Phi_{\text{LLM}}(\mathbf{Q}_k \circ \mathbf{S}_t)
\end{equation}
where $\circ$ denotes token-level concatenation and $\Phi_{\text{LLM}}$ is the frozen LLM. We then compute an unsupervised loss
\begin{equation}
    \mathcal{L}_{\text{total}}
    =
    \sum_{t=1}^N \mathcal{H}\big(a_t^{(k)}\big)
    +
    \lambda \left| \sum_{t=1}^N a_t^{(k)} - \mu \right|
\end{equation}
with $\mathcal{H}(p) = - p \log p - (1-p)\log(1-p)$ the binary entropy, $\mu$ a target anomaly mass reflecting the prior rarity of abnormal events, and $\lambda$ a sparsity weight. The entropy term encourages low-entropy (i.e., confident) anomaly estimates, while the sparsity term regularises the overall anomaly rate without requiring any labels.
The prompt is updated via gradient descent on this unsupervised objective $    \mathbf{Q}_{k+1} = \mathbf{Q}_k - \eta \nabla_{\mathbf{Q}_k}\mathcal{L}_{\text{total}}$
yielding a video-specific, self-consistent prompt $\mathbf{Q}^*$ after a small number of steps. Since only $\mathbf{Q}$ is adjusted and all model parameters remain fixed, this procedure constitutes lightweight test-time adaptation rather than training. Finally, the optimised prompt is used to compute the definitive anomaly scores $a_t = \Phi_{\text{LLM}}(\mathbf{Q}^* \circ \mathbf{S}_t)$.
These scores jointly benefit from (i) the hierarchical, cross-modal structure encoded by hyperbolic fusion, (ii) the LLM's semantic reasoning over structured scene narratives, and (iii) the unsupervised, sparsity-aware adaptation of the prompt.

\subsection{Mahalanobis-Based Cross-Modal Score Refinement}

To anchor the LLM (LLaMA-3B)- derived anomaly scores $a_t$ in the visual stream and mitigate false positives, we introduce a final cross-modal refinement stage. We first obtain visual $E_v(V_t)$ and textual $E_t(S_t)$ embeddings for all clips and summaries using a frozen ImageBind model~\cite{b63}. Central to our approach is the use of Mahalanobis distance to measure the statistical consistency between a summary and the video's global visual feature distribution $\mathcal{V}$. Unlike cosine similarity, which is oblivious to feature distribution, this metric accounts for feature covariance, making it highly sensitive to subtle cross-modal outliers. The distance for summary $S_t$ is:
\begin{equation}
    D_M(S_t, \mathcal{V}) = \sqrt{(E_t(S_t) - \boldsymbol{\mu}_v)^\top \boldsymbol{\Sigma}_v^{-1} (E_t(S_t) - \boldsymbol{\mu}_v)}
    \label{eq:mahalanobis}
\end{equation}
where $\boldsymbol{\mu}_v$ and $\boldsymbol{\Sigma}_v$ are the mean and covariance of the visual distribution $\mathcal{V}$.

We then refine each initial score $a_t$ via weighted aggregation over its $K$ nearest textual neighbors. The influence of each neighbor $a_j$ is weighted by its visual plausibility, $w_j = \exp(-D_M(S_j, \mathcal{V}))$, ensuring that only visually-grounded neighbors contribute significantly:
\begin{equation}
    a'_t = \frac{ \sum_{j \in \mathcal{K}_t} w_j \cdot a_j }{ \sum_{j \in \mathcal{K}_t} w_j }
    \label{eq:score_refinement}
\end{equation}

\section{Experiments and Results} \label{sec:setup}

We evaluate MM-VAD on four key benchmarks dataset: XD-Violence \cite{b3}, UCF-Crime \cite{b2}, ShanghaiTech \cite{liu2018ano_pred}, and UCSD Ped2 \cite{li2013anomaly}.
MM-VAD's performance is measured with frame-level AUC-ROC to differentiate normal and abnormal frames, while also reporting Average Precision (AP) on the XD-Violence benchmark. Our experimental validation is threefold; we first present quantitative comparisons against state-of-the-art methods, followed by qualitative visualisations of anomaly detection. Finally, we conduct a series of ablation studies to systematically analyse the contribution of each component in our model. All experiments are conducted on a single NVIDIA GPU RTX 4080. Detailed hyperparameter settings are provided in \emph{Appendix C} of the supplementary material.

\subsection{Quantitative Results}

We present a comprehensive evaluation of MM-VAD against state-of-the-art methods on the UCF-Crime and XD-Violence benchmarks, with results detailed in Table~\ref{tab:ucf_xd_combined_grouped}. Our core finding is that MM-VAD establishes a new state-of-the-art for training-free VAD, while also demonstrating highly competitive performance against methods that require extensive in-domain training.
MM-VAD consistently outperforms training-free methods by a significant margin. On UCF-Crime, it achieves a top AUC of \textbf{83.24\%}, surpassing the previous best training-free model, MCANet (82.47\%). The advantage is more pronounced on the complex XD-Violence benchmark, where MM-VAD attains an AUC of \textbf{90.03\%} and an AP of \textbf{65.30\%}, outperforming the best method, EventVAD, by \textbf{+2.52\%} AUC and \textbf{+1.26\%} AP. Notably, MM-VAD achieves substantial gains over other prominent training-free baselines, including AUC improvements of \textbf{+1.59\%} and \textbf{+6.51\%} over Flashback-IB on UCF-Crime and XD-Violence, and \textbf{+2.96\%} and  \textbf{+4.67\%} over LAVAD on the same benchmarks. Crucially, our model's performance transcends the training-free category. It exceeds unsupervised methods; for instance, it outperforms C2FPL by \textbf{+2.39\%} AUC on UCF-Crime and by a substantial \textbf{+9.94\%} AUC on XD-Violence. Furthermore, MM-VAD achieves a lead over all one-class baselines, outperforming the best-performing, GODS, by more than\textbf{12\%} AUC on both datasets. 

\begin{table}[t]
\caption{Performance comparison on the UCF-Crime and XD-Violence benchmarks. Methods are grouped by training paradigm: one-class (\colorbox{yellow!20}{yellow}), unsupervised (\colorbox{green!15}{green}), and training-free (\colorbox{purple!10}{purple}). The best performance in each metric is in \textbf{bold}.}
\centering
\small
\setlength\tabcolsep{10pt}
\renewcommand{\arraystretch}{1}
\begin{adjustbox}{max width=1.0\linewidth,center}
\begin{tabular}{l c | c | c c}
\toprule
\textbf{Method} & \textbf{Backbone} 
& \multicolumn{1}{c|}{\textbf{UCF-Crime}} 
& \multicolumn{2}{c}{\textbf{XD-Violence}} \\
\cmidrule(lr){3-3} \cmidrule(lr){4-5}
 & & AUC(\%) & AP(\%) & AUC(\%) \\
\midrule

\rowcolor{yellow!20} LU et al.~\cite{b65} & Dictionary & - & - & 53.56 \\
\rowcolor{yellow!20} SACR~\cite{b66} & - & 72.70 & - & - \\
\rowcolor{yellow!20} BODS~\cite{b26} & I3D-RGB & 68.26 & - & 57.32\\
\rowcolor{yellow!20} GODS~\cite{b26} & I3D-RGB & 70.46 & - & 61.56 \\ 
\rowcolor{green!15} GCL~\cite{b4} & ResNext & 74.20 & - & - \\
\rowcolor{green!15} Tur et al~\cite{b55} & ResNet & 66.85 & - & - \\
\rowcolor{green!15} Tur et al~\cite{b56} & ResNet & 65.22 & - & - \\
\rowcolor{green!15} RareAnom~\cite{b58} & 13D-RGB & - & - & 68.33 \\
\rowcolor{green!15} DYANENT~\cite{b48} & I3D & 79.76 & - & - \\
\rowcolor{green!15} C2FPL~\cite{b67} & 13D & 80.65 & - & 80.09 \\
\rowcolor{purple!10} Zero-Shot CLIP~\cite{b47} & ViT & 53.16 & 17.83 & 38.21 \\
\rowcolor{purple!10} ZS IMAGEBIND (IMAGE) \cite{b63}  & 13D & 53.65 & 27.25 & 58.81 \\
\rowcolor{purple!10} ZS IMAGEBIND (VIDEO) \cite{b63}  & 13D & 55.78 & - & 55.06 \\
\rowcolor{purple!10} LLAVA-1.5~\cite{b69} & ViT & 72.84 & 50.26 & 79.62 \\
\rowcolor{purple!10} Video-LLama2~\cite{zhang2023video} & ViT & 74.42 & 53.57 & 80.21 \\
\rowcolor{purple!10} LAVAD~\cite{b1} & ViT & 80.28 & 62.01 & 85.36 \\
\rowcolor{purple!10} AnyAnomaly~\cite{ahn2025anyanomaly} & ViT & 80.70 & - & - \\
\rowcolor{purple!10} Flashback-IB~\cite{lee2025} & ViT & 81.65 & 60.13 & 83.52 \\
\rowcolor{purple!10} MCANet~\cite{dev2024mcanet} & ViT & 82.47 & - & 87.43 \\
\rowcolor{purple!10} EventVAD~\cite{event} & ViT & 82.03 & 64.04 & 87.51 \\
\rowcolor{purple!10} \textbf{MM-VAD} & ViT & \textbf{83.24} & \textbf{65.30} & \textbf{90.03} \\

\bottomrule
\end{tabular}
\end{adjustbox}
\label{tab:ucf_xd_combined_grouped}
\end{table}

To evaluate the \emph{modality-agnostic} behaviour of MM-VAD, we further test the framework on the video-only ShanghaiTech \cite{liu2018ano_pred} and UCSD Ped2 \cite{li2013anomaly} datasets. These benchmarks, along with the vision-only UCF-Crime results, allow us to verify that the method remains effective without audio input. As shown in Table \ref{tab:lavad_ours_sh_ucsd}, MM-VAD outperforms LAVAD by \textbf{+2.40\%} AUC points on ShanghaiTech and by \textbf{+2.70\%} points on UCSD Ped2, demonstrating consistent gains across all video-only settings. Due to space constraints, only the primary comparison is reported in the main paper, while extended results under additional learning paradigms are provided in the \emph{Appendix~A} of the supplementary material. 
\begin{table}[htbp]
\centering

\caption{Comparison of frame-level AUC (\%) for training-free methods on ShanghaiTech and UCSD Ped2.}
\label{tab:lavad_ours_sh_ucsd}
\setlength{\tabcolsep}{15pt}
\renewcommand{\arraystretch}{0.1}
\small
\begin{tabular}{lcc}
\toprule
\textbf{Dataset} & \textbf{LAVAD~\cite{b1}} & \textbf{MM-VAD} \\
\midrule
ShanghaiTech \cite{liu2018ano_pred} & 94.55 & \textbf{96.95} \\
UCSD Ped2 \cite{li2013anomaly}    & 96.11 & \textbf{98.81} \\
\bottomrule
\end{tabular}
\end{table}

\subsection{Qualitative Results}

\begin{figure}[htbp]
    \centering
    \includegraphics[width=0.99\textwidth]{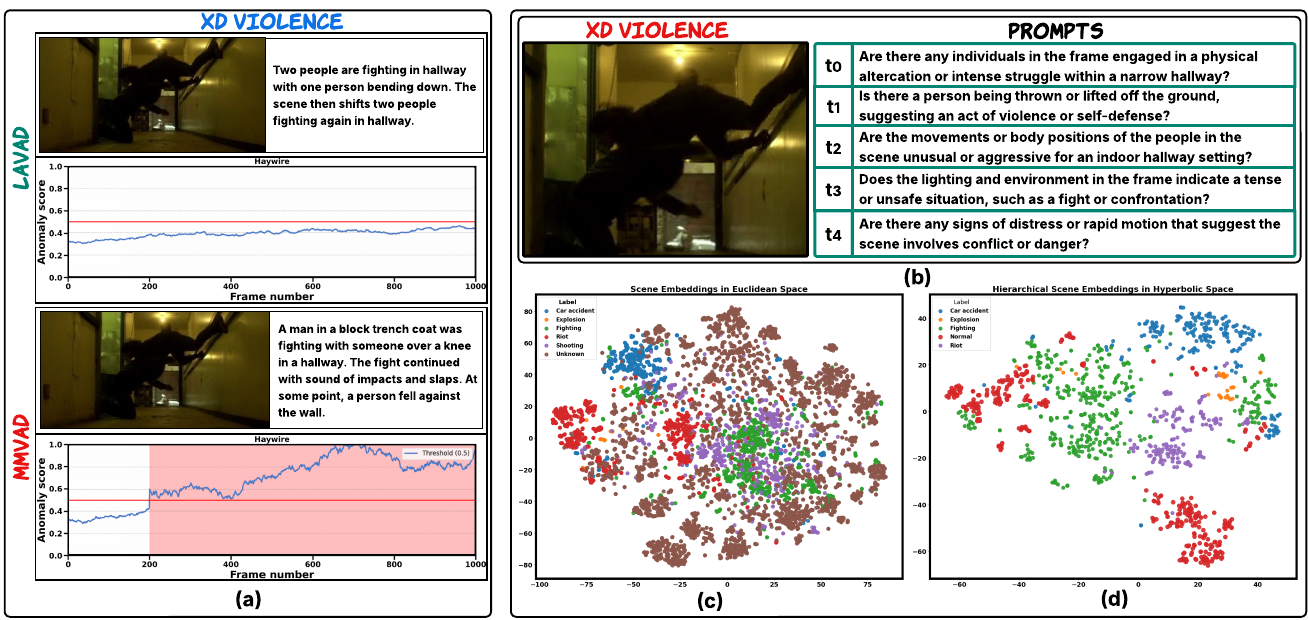}
   \caption{Overview of MM-VAD on XD-Violence. (a) MM-VAD outperforms LAVAD with sharper anomaly localisation and richer action-aware descriptions. (b) Adaptive prompts ($t_0$--$t_4$) for frozen-LLM querying. (c) Euclidean embeddings. (d) Hyperbolic embeddings with clearer structure and better separation.
   }
\label{fig:XD_Ques}
\end{figure}

To complement our quantitative results, we provide qualitative visualisations in Figure~\ref{fig:XD_Ques}a that compare MM-VAD against LAVAD on challenging examples from XD-Violence. These examples illustrate our model's superior ability to not only accurately detect anomalies but also to generate semantically rich and contextually relevant textual \emph{explanations} for its decisions. On the XD-Violence example, a clear fighting scene unfolds. LAVAD fails to recognize the anomaly, with its score remaining consistently below the detection threshold. Its textual summary is generic, vaguely describing "two people fighting". In stark contrast, MM-VAD correctly identifies the entire anomalous segment with a high, sustained anomaly score. More importantly, its textual description is remarkably detailed and multimodal, capturing specific actions ("fighting with someone over a knee," "a person fell against the wall") and even auditory cues ("with the sound of impacts and slaps"), demonstrating a deep \emph{explainable}, grounded understanding of the event.

As illustrated in Figure~\ref{fig:XD_Ques}b, the adaptive query optimiser progressively refines its questions from broad, low-risk queries in clearly normal segments to more pointed, violence-oriented prompts as the hallway becomes crowded and interactions grow ambiguous. This highlights its ability to allocate more specialised reasoning to segments that are both visually uncertain and semantically high-risk, rather than treating all clips with a fixed set of hand-crafted questions. Figure \ref{fig:XD_Ques}c and \ref{fig:XD_Ques}d show 2D t-SNE visualisations of high-dimensional scene embeddings.  The baseline model (a) produce embeddings with significant overlap, particularly between semantically anomalies like 'Riot' (red) and 'Fighting' (green). Critically, non-anomalous data is relegated to a diffuse, space-filling 'Unknown' category (brown), indicating the model lacks a coherent representation of normalcy. In contrast, the hyperbolic model (b), yields substantially more compact and well-separated clusters. Anomalies are more distinct, and most importantly, the scattered 'Unknown' points are replaced by a tight, centrally-located 'Normal' class (red). This demonstrates that hyperbolic geometry provides a superior inductive bias for learning a structured manifold of scenes, separating diverse anomalies from a well-defined concept of normalcy. Additional qualitative results are presented in \emph{Appendix B} of the supplementary material.

\subsection{Ablation Study}
\label{sec:ablation_study}
\subsubsection{Component-wise Analysis}
We conduct an ablation study to quantify the contribution of each module in MM-VAD, with results summarised in Table~\ref{tab:ablation}. On \textbf{XD-Violence}, the baseline using visual captions $C^{\text{vis}}$, audio captions $C^{\text{aud}}$, and Mahalanobis refinement $D_M$ achieves 74.56\% AUC. Cleaning the visual captions $C'^{\text{vis}}$ improves performance to 80.05\% (+5.49), and replacing Euclidean fusion with hyperbolic fusion $z_t^{(\mathcal{F})}$ yields 81.97\% (+7.41), demonstrating the importance of caption quality and non-Euclidean fusion. Combining these components in a vision-only setting reaches 85.53\%, and adding audio further raises the score to 87.83\%. Enabling the adaptive question optimiser $Q_{t+1}$ provides the final improvement, achieving 90.03\% AUC. On \textbf{UCF-Crime}, which is video-only, the same progression increases the baseline from 70.10\% to 80.41\% with caption cleaning and hyperbolic fusion, and $Q_{t+1}$ delivers the final boost to 83.24\%. Overall, the ablation shows that each module contributes meaningfully and that their combination is necessary to attain the full performance of MM-VAD.
\newcommand{\cmark}{\ding{51}} 
\newcommand{\xmark}{\ding{55}} 
\begin{table}[htbp]
\small
\centering
\caption{Ablation study of MM-VAD's components}
\label{tab:ablation}
\setlength{\tabcolsep}{10pt}  
\renewcommand{\arraystretch}{0.8}  
\resizebox{\columnwidth}{!}{%
\begin{tabular}{cccccc|cc}
\toprule
\textbf{$C^{\text{vis}}$} & \textbf{$C^{\text{aud}}$} & \textbf{$C'^{\text{vis}}$} & \textbf{$z_t^{(\mathcal{F})}$} & \textbf{$D_M$} & \textbf{$Q_{t+1}$} & \textbf{XD-Violence} & \textbf{UCF-Crime} \\
\cmidrule(lr){7-8} 
\multicolumn{6}{c|}{} & \textbf{AUC (\%)} & \textbf{AUC (\%)} \\
\midrule
\cmark & \cmark & \xmark & \xmark & \cmark & \xmark & 74.56 & 70.10 \\
\cmark & \cmark & \cmark & \xmark & \cmark & \xmark & 80.05 & 77.80 \\
\cmark & \cmark & \xmark & \cmark & \cmark & \xmark & 81.97 & 78.87 \\
\cmark & \xmark & \xmark & \cmark & \cmark & \cmark & 83.24 & 79.75 \\
\cmark & \xmark & \cmark & \cmark & \cmark & \xmark & 85.53 & 80.41 \\
\cmark & \cmark & \cmark & \cmark & \cmark & \xmark & 87.83 & 80.41 \\
\textbf{\cmark} & \textbf{\cmark} & \textbf{\cmark} & \textbf{\cmark} & \textbf{\cmark} & \textbf{\cmark} & \textbf{90.03} & \textbf{83.24} \\
\bottomrule
\end{tabular}%
}
\end{table}

\subsubsection{Backbone Analysis: VLM and LLM Comparisons}
Table~\ref{tab:combined_ablation} summarises controlled backbone ablations across vision--language and language model choices. In Table~\ref{tab:main_cross_comb}, replacing Blip-2 with VLRM consistently improves performance for both LAVAD and MM-VAD, while substituting LLaMA-2-13B (L2-13B) with LLaMA-3B (L-3B) yields a further but smaller gain. Across all shared component settings, MM-VAD remains superior to LAVAD, indicating that the proposed framework benefits more effectively from stronger backbone combinations. The best cross-combination is obtained with VLRM and L-3B, where MM-VAD reaches 83.24\% AUC on UCF-Crime and 90.03\% on XD-Violence. Table~\ref{tab:llm_ablation} further isolates the effect of the language model with VLRM fixed. Although all evaluated LLMs provide competitive results, performance improves progressively from DeepSeek (DS), Qwen-4B (Q-4B), Qwen-30B (Q-30B), and Mistral-24B (M-24B) variants to L2-13B, with L-3B achieving the strongest overall performance. These findings jointly validate VLRM and L-3B as the most effective backbone pair for the final framework.
\begin{table}[htbp]
\caption{Controlled component and model ablations. (a) Cross-combination study between LAVAD and MM-VAD under different captioning and LLM backbones. (b) LLM ablation with VLRM fixed to isolate the effect of the language model.}
\label{tab:combined_ablation}
\renewcommand{\arraystretch}{1.0}
\small

\begin{subtable}[t]{\linewidth}
\raggedright
\caption{Cross-combination study between LAVAD and MM-VAD under shared component settings.}
\label{tab:main_cross_comb}
\setlength{\tabcolsep}{6pt}
\resizebox{\linewidth}{!}{%
\begin{tabular}{@{}cccc|cc|cc@{}}
\toprule
\multicolumn{4}{c|}{\textbf{Shared Components}} & \multicolumn{2}{c|}{\textbf{LAVAD ~\cite{b1}}} & \multicolumn{2}{c@{}}{\textbf{MM-VAD}} \\
\cmidrule(r){1-4} \cmidrule(lr){5-6} \cmidrule(l){7-8}
\textbf{BLIP-2} & \textbf{VLRM} & \textbf{L2-13B} & \textbf{L-3B}
& \textbf{UCF-Crime} & \textbf{XD-Violence}
& \textbf{UCF-Crime} & \textbf{XD-Violence} \\
\cmidrule(lr){5-6} \cmidrule(l){7-8}
& & & & \textbf{AUC (\%)} & \textbf{AUC (\%)} & \textbf{AUC (\%)} & \textbf{AUC (\%)} \\
\midrule
$\checkmark$ & $\times$      & $\checkmark$ & $\times$      & 80.28 & 85.36 & 82.68 & 89.11 \\
$\checkmark$ & $\times$      & $\times$     & $\checkmark$  & 80.38 & 85.41 & 82.81 & 89.33 \\
$\times$     & $\checkmark$  & $\checkmark$ & $\times$      & 80.95 & 86.01 & 83.13 & 89.97 \\
$\times$     & $\checkmark$  & $\times$     & $\checkmark$  & 81.04 & 86.27 & \textbf{83.24} & \textbf{90.03} \\
\bottomrule
\end{tabular}%
}
\end{subtable}

\vspace{0pt}

\begin{subtable}[t]{\linewidth}
\centering
\caption{LLM ablation with VLRM fixed.}
\label{tab:llm_ablation}
\setlength{\tabcolsep}{6pt}
\makebox[\linewidth][c]{%
\resizebox{0.9\linewidth}{!}{%
\begin{tabular}{@{}cccccc|c|c@{}}
\toprule
\multicolumn{6}{c|}{\textbf{LLMs}} & \multicolumn{2}{c@{}}{\textbf{Datasets}} \\
\cmidrule(r){1-6} \cmidrule(l){7-8}
\textbf{DS} & \textbf{Q-4B} & \textbf{Q-30B} & \textbf{M-24B} & \textbf{L2-13B} & \textbf{L-3B} & \textbf{UCF-Crime} & \textbf{XD-Violence} \\
\cmidrule(l){7-8}
& & & & & & \textbf{AUC (\%)} & \textbf{AUC (\%)} \\
\midrule
$\checkmark$ & $\times$     & $\times$     & $\times$     & $\times$     & $\times$     & 79.44          & 85.44 \\
$\times$     & $\checkmark$ & $\times$     & $\times$     & $\times$     & $\times$     & 79.55          & 86.40 \\
$\times$     & $\times$     & $\checkmark$ & $\times$     & $\times$     & $\times$     & 80.15          & 87.21 \\
$\times$     & $\times$     & $\times$     & $\checkmark$ & $\times$     & $\times$     & 81.05          & 87.95 \\
$\times$     & $\times$     & $\times$     & $\times$     & $\checkmark$ & $\times$     & 83.13          & 89.97 \\
$\times$     & $\times$     & $\times$     & $\times$     & $\times$     & $\checkmark$ & \textbf{83.24} & \textbf{90.03} \\
\bottomrule
\end{tabular}%
}%
}
\end{subtable}

\end{table}

Importantly, when the VLM and LLM backbones are fixed to the same configuration used by LAVAD, MM-VAD still achieves consistent performance improvements. This confirms that the observed gains arise from the proposed geometry-aware reasoning framework rather than from stronger captioning or language models.
We present \textbf{additional ablation studies} in \emph{Appendix D} of the supplementary material, including an analysis of different similarity metrics, configurations for hyperbolic space (e.g., distances and curvature),the optimal value of K for our score refinement module, and learnable questions prompts examples using adaptive query optimisation.


\subsubsection{Time Complexity Analysis}
MM-VAD is not only more accurate but also more efficient in practice. As shown in Table~\ref{tab:time_transposed}, our full pipeline processes a batch of 16 frames in 27.0\,s, compared to 32.5\,s for LAVAD, corresponding to 1.69\,s/frame versus 2.03\,s/frame. This yields a 16.8\% speed-up overall. The gain mainly comes from the reasoning stages, where MM-VAD reduces summary generation (9.0\,s vs.\ 13.0\,s) and scoring (4.0\,s vs.\ 7.0\,s). Notably, this improvement is achieved despite including additional audio processing steps (1.3\,s total) that are absent in LAVAD. MM-VAD is also more memory-efficient, requiring 26.5\,GB VRAM compared with 33.5\,GB for LAVAD.

\begin{table*}[htbp]
\centering
\tiny
\caption{Per-step runtime comparison between \textbf{MM-VAD} and \textbf{LAVAD}. Lower values indicate faster processing and are preferred. The best runtime for each step is highlighted in \textbf{bold}. Shortened headers: A-Cap = Audio-caption, I-Cap = Image caption, Clean = Image caption clean, Summ = LLM-based summary generation, Sco = anomaly score, Refine = Score Refinement, and Batch = Time/Batch (16 frames).}
\label{tab:time_transposed}
\setlength{\tabcolsep}{3pt}
\renewcommand{\arraystretch}{1.2}
\begin{tabular}{lcccccccccccc}
\toprule
\textbf{Method} & \textbf{Frame} & \textbf{Audio} & \textbf{A-Cap} & \textbf{I-Cap} & \textbf{Idx$_1$} & \textbf{Clean} & \textbf{Summ} & \textbf{Sco} & \textbf{Idx$_2$} & \textbf{Refine} & \textbf{Batch} & \textbf{Sec/Frame} \\
\midrule

LAVAD ~\cite{b1} 
& \textbf{0.2} 
& \textbf{0.0} 
& \textbf{0.0} 
& 1.3 
& \textbf{1.0} 
& \textbf{4.0} 
& 13.0 
& 7.0 
& \textbf{1.0} 
& \textbf{5.0} 
& 32.5 
& 2.03 \\
MM-VAD 
& \textbf{0.2} 
& 0.3 
& 1.0 
& \textbf{1.0} 
& \textbf{1.0} 
& \textbf{4.0} 
& \textbf{9.0} 
& \textbf{4.0} 
& \textbf{1.0} 
& 5.5 
& \textbf{27.0} 
& \textbf{1.69} \\
\bottomrule
\end{tabular}
\end{table*}

\section{Conclusion}

MM-VAD reframes video anomaly detection as a multimodal semantic reasoning problem rather than a feature-matching task. Through the combination of hyperbolic fusion and adaptive question–answering, the framework constructs representations that capture the hierarchical structure of visual and auditory events and allows an LLM to reason over these structures in a scene-aware way. This leads to behaviour that differs markedly from existing training-free VAD approaches, since the model generates anomaly scores supported by coherent textual explanations, remains stable when a modality is absent, and preserves discriminative power across diverse anomaly types. 
Across four benchmark datasets, MM-VAD delivers consistently strong performance, surpassing prior zero-shot and several unsupervised methods, with ablations showing that its gains arise from the combined contribution of all modules. The results also indicate that performance is most affected by noisy captions, such as those generated in cluttered scenes or from imperfect audio transcripts, and by anomalies that develop gradually over long durations, which can exceed the capacity of short-window summaries. These observations highlight promising directions for future work involving more robust caption refinement and temporally enriched reasoning modules that can extend MM-VAD’s sensitivity to long-horizon and weak-signal anomaly patterns.

\bibliographystyle{splncs04}
\bibliography{main}
\end{document}